\documentclass[10pt,twocolumn,letterpaper]{article}

\usepackage{cvpr}
\usepackage{times}
\usepackage{epsfig}
\usepackage{graphicx}
\usepackage{amsmath}
\usepackage{amssymb}
\usepackage{epstopdf}
\usepackage{algorithm}  
\usepackage{algorithmic}  
\usepackage{amsmath}
\usepackage{latexsym}
\usepackage{makecell}
\usepackage{multirow,diagbox}
\usepackage{array}
\usepackage{booktabs}

\usepackage[breaklinks=true,bookmarks=false]{hyperref}

\cvprfinalcopy 

\def\cvprPaperID{4946} 

\begin{document}

\title{Tightness-aware Evaluation Protocol for Scene Text Detection}

\author{Yuliang Liu, Lianwen Jin$^{*}$, Zecheng Xie, Canjie Luo, Shuaitao Zhang, Lele Xie\\
College of Electronic Information Engineering, South China University of Technology \\
{\tt\small liu.yuliang@mail.scut.edu.cn; lianwen.jin@gmail.com}
}

\maketitle

\begin{abstract}
   Evaluation protocols play key role in the developmental progress of text detection methods.
   There are strict requirements to ensure that the evaluation methods are fair, objective and reasonable. 
   However, existing metrics exhibit some obvious drawbacks: 
   1) They are not goal-oriented;
   2) they cannot recognize the tightness of detection methods;
   3) existing one-to-many and many-to-one solutions involve inherent loopholes and deficiencies. 
   Therefore, this paper proposes a novel evaluation protocol called  Tightness-aware Intersect-over-Union (TIoU) metric that could quantify completeness of ground truth, compactness of detection, and tightness of matching degree. 
   Specifically, instead of merely using the IoU value, two common detection behaviors are properly considered; meanwhile, directly using the score of TIoU to recognize the tightness. In addition, we further propose a straightforward method to address the annotation granularity issue, which can fairly evaluate word and text-line detections simultaneously.
   By adopting the detection results from published methods and general object detection frameworks, comprehensive experiments on ICDAR 2013 and ICDAR 2015 datasets are conducted to compare recent metrics and the proposed TIoU metric. The comparison demonstrated some promising new prospects, e.g., determining the methods and frameworks for which the detection is tighter and more beneficial to recognize. Our method is extremely simple; however, the novelty is none other than the proposed metric can utilize simplest but reasonable improvements to lead to many interesting and insightful prospects and solving most the issues of the previous metrics.
   The code is publicly available at \url{https://github.com/Yuliang-Liu/TIoU-metric}.
\end{abstract}

\section{Introduction}
\label{sec:intro}
\begin{figure}[htb]
\begin{minipage}[b]{.48\linewidth}
  \centering
  \centerline{\includegraphics[width=4cm]{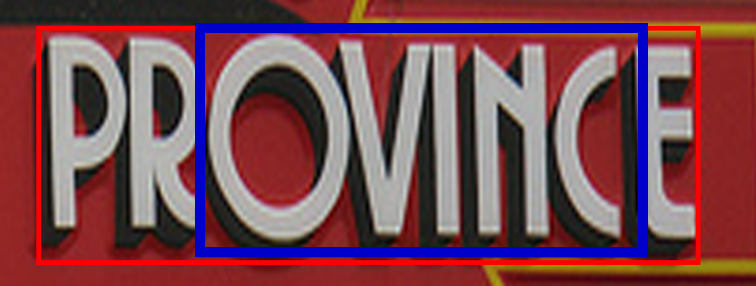}}
  \centerline{(a) Cutting.}\medskip
\end{minipage}
\hfill
\begin{minipage}[b]{0.48\linewidth}
  \centering
  \centerline{\includegraphics[width=4cm]{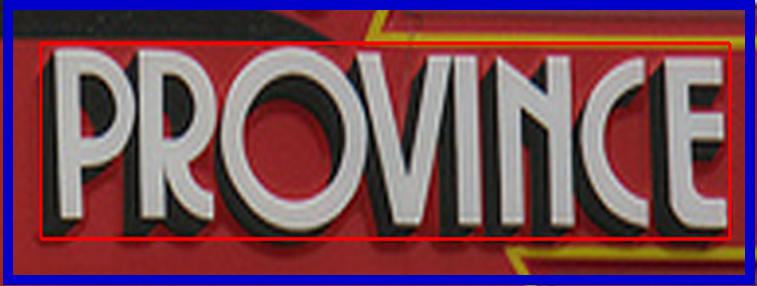}}
  \centerline{(b) Pure.}\medskip
\end{minipage}
\vfill
\begin{minipage}[b]{.48\linewidth}
  \centering
  \centerline{\includegraphics[width=4cm]{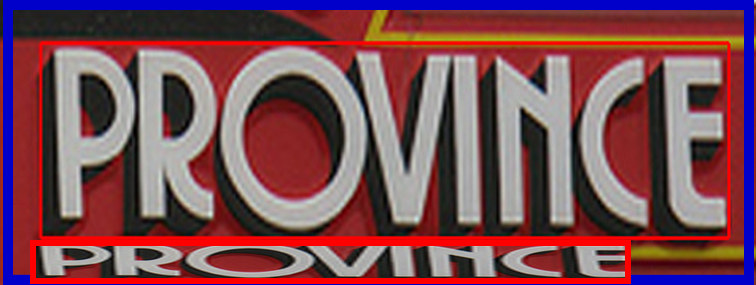}}
  \centerline{(c) Outlier-GTs.}\medskip
\end{minipage}
\hfill
\begin{minipage}[b]{0.48\linewidth}
  \centering
  \centerline{\includegraphics[width=4cm]{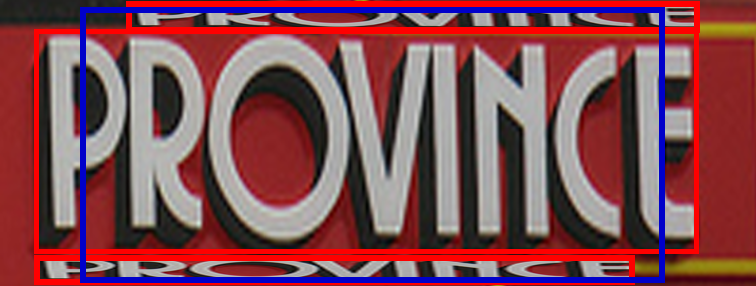}}
  \centerline{(d) Cutting \& Outlier-GTs.}\medskip
\end{minipage}
\caption{Unreasonable cases obtained using recent evaluation metrics. (a), (b), (c), and (d) all have the same IoU of 0.66 against the GT. Red: GT. Blue: detection. }
\label{fig:intro1}
\end{figure}
Recent metrics for evaluating text detection have been adopted from the object detection Pascal VOC metric~\cite{everingham2015pascal}.
 However, unlike object detection, text detection tasks require the bounding box to be tighter because the primary goal of detection is to recognize the text. Simply adopting the same IoU metric for text detection leads to the following issues:
\begin{itemize}
  \item As shown in Fig. \ref{fig:intro1} (a), detection over a fixed IoU threshold with the ground truth (GT) may not completely recall the text (some characters are missed); however, previous metrics consider that the GT has been entirely recalled.
  \item As shown in Figs. \ref{fig:intro1} (b), (c), and (d), detection over a fixed IoU threshold with the GT may still contain background noise; however, previous metrics consider such detection to have 100\% precision.
  \item As shown in Fig. \ref{fig:intro1}, previous metrics consider detections (a), (b), (c), and (d) to be equivalent perfect detections because they all have the same IoU value that is higher than a threshold. However, considering that the primary goal of detection is to recognize the text, these detections are not equivalent: 1) In (a), there is no way to recognize the characters outside the detection bounding box; 2) in (c), it is very difficult for a recognizer to distinguish which is the target GT; 3) the issues pertaining to both (a) and (c) can simultaneously occur for (d);
  4) as for (b), it is easy for a normal text recognizer to recognize the content correctly. 
  \item Previous metrics severely rely on an IoU threshold. However, if a relatively high IoU threshold is set, some satisfactory bounding boxes may be discarded (e.g., if 0.7 is set as the threshold, the detection in Fig. \ref{fig:intro1} (b) will be misjudged); if a low IoU threshold is set, several inexact bounding boxes would be included.
\end{itemize}

The essential reasons that cause such inequalities are: 1) The detections in Figs. \ref{fig:intro1} (a) and (d) cut the GT region; 2) the detections in Figs. \ref{fig:intro1} (c) and (d) both contain outlier-GT; and 3) recent metrics use binary results (0 or 100\%) to represent the final recall or precision score.

To solve these issues, an intuitive solution is to penalize the exceptional detections. 
In the proposed Tightness-aware Intersection-over-Union (TIoU) metric, we use the occupation ratio of detection to GT and occupation ratio of outlier-GT to detection as penalty factors, respectively. 
In addition, we directly use the score of the TIoU as the value of recall and precision, and thus, the compact degree among different methods can be distinguished. The TIoU metric targets the following three characteristics, which are also the main contributions of this work:
\begin{itemize}
  \item {\bf {Completeness.}} Using the TIoU metric would force methods to pay more attention to recalling every part of the GT, i.e., ensuring the completeness of GT.
  \item {\bf {Compactness.}} Because the detections of outlier-GT will be punished by TIoU, the compactness of the detection would receive more attention.
  \item {\bf {Tightness-aware.}} TIoU can distinguish the tightness among different detection methods, i.e., a 0.9 IoU detection would be much better than a 0.5 IoU detection in our metric.
\end{itemize}

\section{Related work}
Evaluation methods \cite{mariano2002performance,wolf2006object,everingham2015pascal} have been an important research topic for several decades. For scene text detection, there are four  mainstream evaluation methods that are largely identical but have minor differences: ICDAR 2003 (IC03) \cite{lucas2003icdar}, ICDAR 2013 (IC13) \cite{Karatzas2013ICDAR,wolf2006object}, ICDAR 2015 (IC15) \cite{karatzas2015icdar}, and AP-based methods \cite{everingham2015pascal,Shi2017ICDAR2017}. 

\subsection{ICDAR 2003 Evaluating Method}
The early IC03 metric \cite{lucas2003icdar} is based on the notions used by the information retrieval community to calculate precision and recall, which are as follows:
\begin{equation}\label{eq:ic03Recall}
   Recall(G,D) = \frac{\sum_{i=1}^{|G|}BestMatch_{G}(G_{i})}{|G|},
\end{equation}
\begin{equation}\label{eq:ic03Precision}
   Precision(G,D) = \frac{\sum_{j=1}^{|D|}BestMatch_{D}(D_{j})}{|D|},
\end{equation}
where, $BestMatch_{G}$ and $BestMatch_D$ indicate the result of the closest match between detection and ground truth rectangles, as defined below \cite{wolf2006object}:
\begin{equation}\label{eq:ic03RecallBestMatch}
   BestMatch_{G}(G_{i}) = \max_{j = 1...|D|}\frac{2\cdot Area(G_i\cap D_j)}{Area(G_i)+Area(D_j)},
\end{equation}
\begin{equation}\label{eq:ic03PrecisionBestMatch}
   BestMatch_{D}(D_{j}) = \max_{i = 1...|G|}\frac{2\cdot Area(D_j\cap G_i)}{Area(D_j)+Area(G_i)}.
\end{equation}
The matching mechanism involves finding the perfect-matching pairs to detect the matching value, i.e., if a rectangle is perfectly matched, the match value is unity, else the value is less than 1. The disadvantages of the IC03 metric are as follows: 1) Multiple detections can be repeatedly matched to the same GT;
2) as the authors reported in \cite{lucas2003icdar} themselves, this evaluation scheme only considers one-to-one (OO) matches, and thus one-to-many (OM) and many-to-one (MO) matches \cite{wolf2006object} are considered to have zero recall and precision. However, practically, word-level and text-line detections can both be conducive to recognition.

\subsection{ICDAR2013 Evaluation Method}
IC13~\cite{Karatzas2013ICDAR,wolf2006object} has three evaluation metrics: IC13, DetEval, and IoU.  The IoU metric is subsequently adopted following the IC15 \cite{karatzas2015icdar} metric, which is discussed in the next subsection.

\subsubsection{ICDAR2013, DetEval}
Unlike Eqs. \ref{eq:ic03RecallBestMatch} and \ref{eq:ic03PrecisionBestMatch}, the criteria of these two evaluations are based on mutual overlap rates between detection ({$\{D_j\}_j$}) and ground truth ($\{G_i\}_i$): 
\begin{equation}\label{eq:13precisionTP}
   \frac{A(G_i\cap D_j)}{A(D_j)} > tp,
\end{equation}
\begin{equation}\label{eq:13precisionTR}
   \frac{A(G_i\cap D_j)}{A(G_i)} > tr, 
\end{equation}
where, $tp$ and $tr$ are the thresholds of precision and recall, respectively. The metrics evaluate the methods using three steps including OO, OM, and MO. Unlike the IC03 metric, OO guarantees that each GT can be at most matched once, and multiple matching detections are considered false positives. 

{\bf {OM}} indicates that a GT is matched by a set of detection results, and it should satisfy two requirements: a) Sufficient detections covering the GT; b) each contributing detection is covered enough by the GT. If these conditions are satisfied, the precision value of each detection box and recall value of the GT are both set to 0.8 \cite{Karatzas2013ICDAR}. 

{\bf {MO}} indicates that a detection is matched by a set of GT. The following two requirements must be satisfied: a) The detection must contain sufficiently overlapping GT; b) each GT must be recalled with sufficient area. If these conditions are satisfied, the recall value of each GT and the precision value of this detection are both set to 1.

\begin{figure}[htb]
\centering\centerline{\includegraphics[width = 7.8cm, height = 3.7cm]{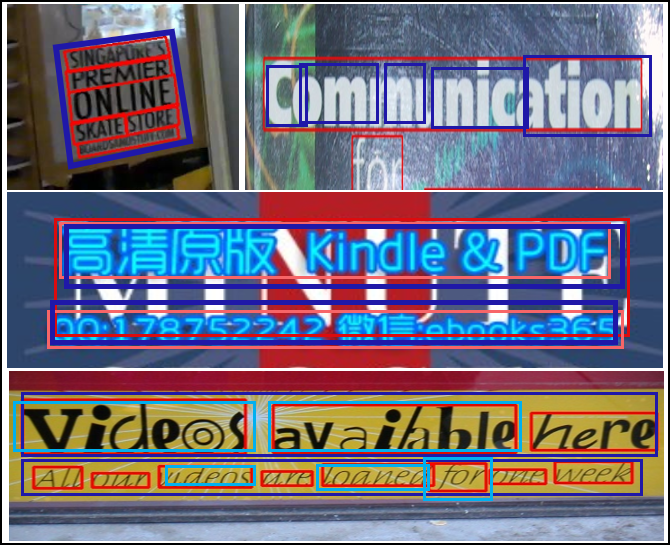}}
\caption{The first row shows examples of many-to-one and one-to-many matching. The second and third rows show some failure matching cases (using IC13 metric) because of the matching order. Red: GT bounding box. Blue: detections. Jacinth: GT bounding boxes that can not be recalled because of one-to-many matching. Bluish: detection box that causes failure of many-to-one matching.}\label{fig:one-many-and-failure}
\end{figure}

Although these two matching methods can overcome the problem of inconsistency of the annotation granularity to some extent, such metrics still involve some unsatisfactory circumstances (examples are shown in Fig. ~\ref{fig:one-many-and-failure}):
\begin{itemize}
   \item For MO matching, a rough detection can recall all GTs, as shown in the left image of the first row of Fig. \ref{fig:one-many-and-failure}. Such rough detection may not be proper recognized, which is relatively unfair for those methods that can granularly detect the text. 
   \item For OM matching, many segmented detections are considered as correct results, which is unfair for singularly perfect detection. Some examples are shown in the right image of the first row of Fig. \ref{fig:one-many-and-failure}. It is worth mentioning that although 0.8 is set to penalize over-segmentation, loophole still exists. For example, if the overall precision is less than 0.8, a method that separates a perfect detection into numerous small over-segmented OM detections (e.g., 20) can make the precision close to 0.8. An example is given below:
\begin{equation}\label{eq:one-to-many_ori}
   origin\_precision = \frac{0+1+0+0}{4} = 0.25
\end{equation}
\begin{equation}\label{eq:one-to-many_cheat}
   fake\_precision = \frac{0+\overbrace{0.8+...+0.8}^{20}+0+0}{23} = 0.7
\end{equation}

   \item For the IC13 metric, {\bf {the matching order}} is OO, OM and MO. Because each GT and detection can only be matched once, the OO matching pairs would not be calculated during OM or MO matching, which would affect the detection results. As shown in the bottom image of Fig. \ref{fig:one-many-and-failure}, the long detections do not satisfy MO matching because the bluish detections have already OO matched with some GT in advance. Thus, the long detections are regarded as false positives and the rest of the GTs are considered to be not recalled.

   \item For DetEval evaluation, OM and MO are evaluated in advance before OO, which may lead to unsatisfactory results. As shown in the middle image of Fig. \ref{fig:one-many-and-failure}, because OM is validated in advance, the two detection bounding boxes have already been used. Thus, the two jacinth GTs are considered not to be recalled. 
\end{itemize}


\subsection{ICDAR 2015 IoU Metric}
The IC15 metric \cite{karatzas2015icdar} follows the same metric as Pascal VOC~\cite{everingham2015pascal}. Under this metric, detections are assigned to ground truth objects and judged to be true or false positives by measuring the bounding box overlap. To be considered a correct detection, the value of Intersection-over-Union (IoU, defined in Eq. \ref{eq:15IoU}) must exceed 0.5.
\begin{equation}\label{eq:15IoU}
   \frac{A(G_j\cap D_i)}{A(G_j\cup D_i)} > 0.5.
\end{equation}

Detections are assigned to true positives only when they satisfy the overlap criterion and they have the top ranking confidence to the target GT. Ground truth objects with no matching detections are false negatives. Although this metric has received the most attention, it still has many drawbacks, which have been pointed out in the introduction section.

\subsection{AP-based evaluation method}
To avoid finetuning the output detection confidence, dataset such as RCTW-17~\cite{Shi2017ICDAR2017} has adopted interpolated average precision as the main detection evaluation metric: For a given task and class, the precision-recall curve is computed based on the method's ranked output. Basically, this metric relies on the IoU metric to calculate the precision and recall in advance.

\section{Methodology}

\begin{figure*}[htb]
\begin{minipage}[b]{.48\linewidth}
  \centering
  \centerline{\includegraphics[width=9cm, height = 5.0cm]{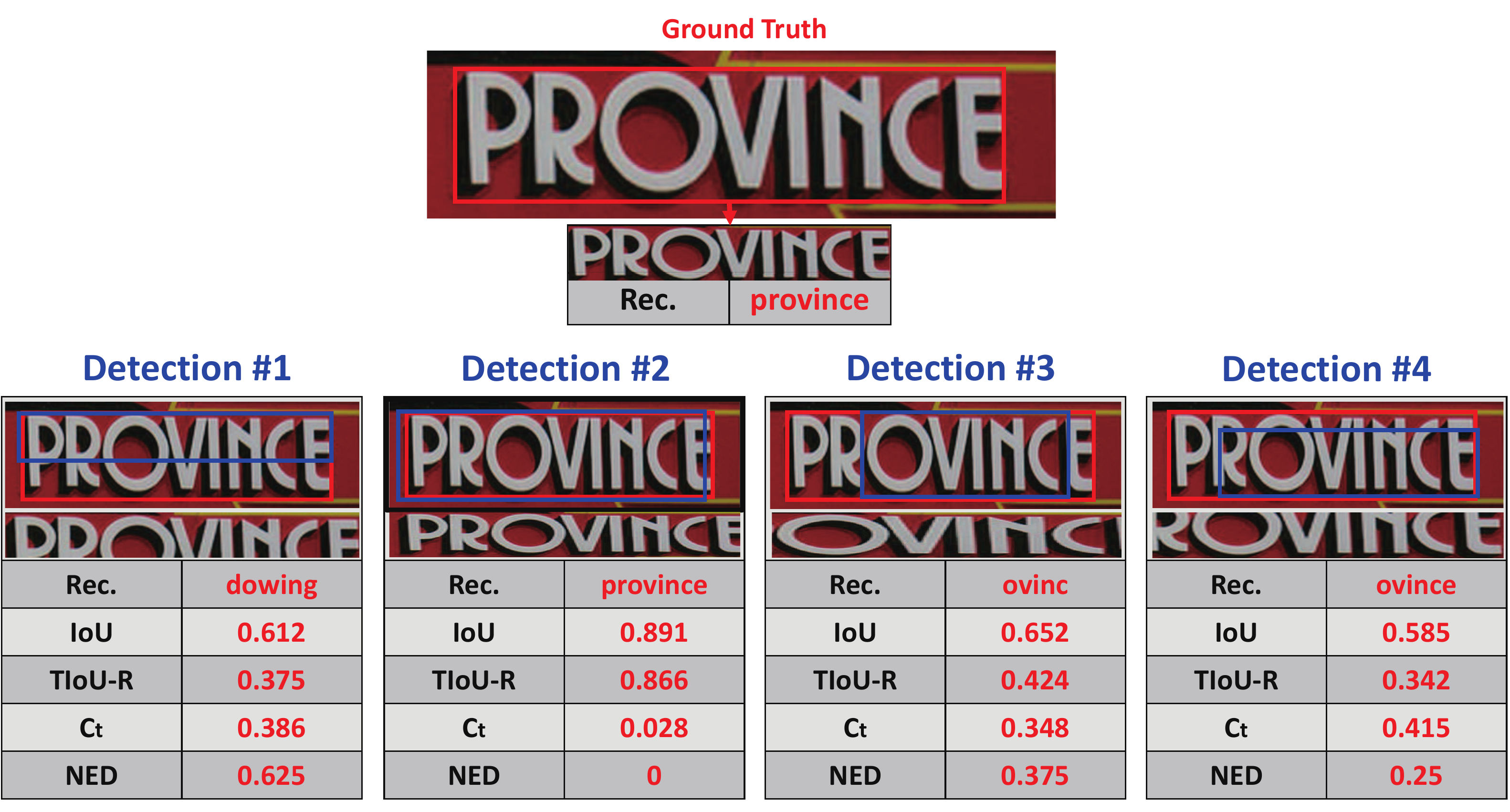}}
  \centerline{(a) Cutting effect.}\medskip
\end{minipage}
\hfill
\begin{minipage}[b]{0.48\linewidth}
  \centering
  \centerline{\includegraphics[width=9cm, height = 5.0cm]{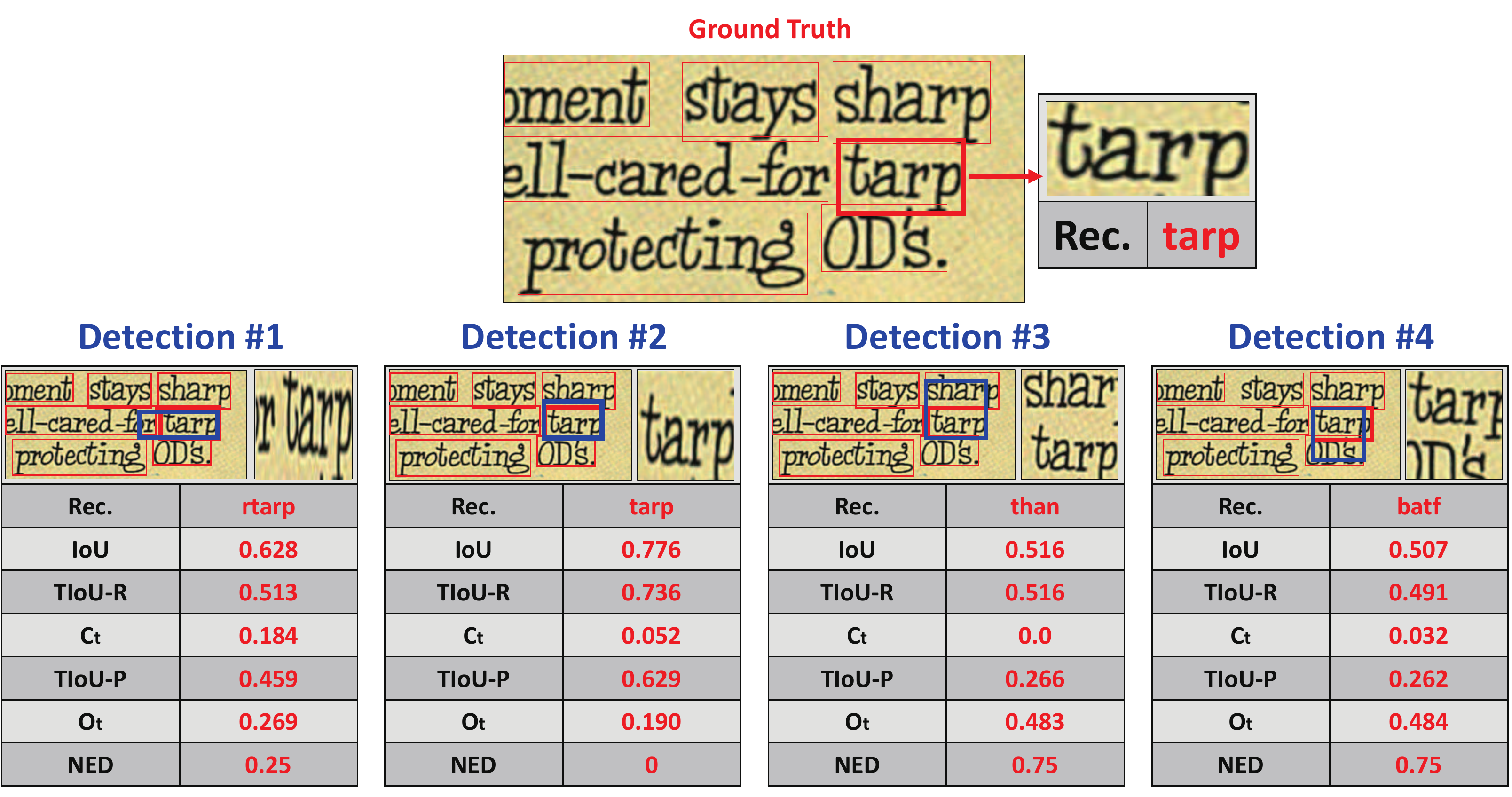}}
  \centerline{(b) Outlier effect.}\medskip
\end{minipage}
\caption{Qualitative visualization of TIoU metric. Blue: Detection. Bold red: Target GT region. Light red: Other GT regions. Rec.: Recognition results by CRNN \cite{shi2017end}. NED: Normalized edit distance. Previous metrics evaluate all detection results and target GTs as 100\% precision and recall, respectively, while in TIoU metric, all matching pairs are penalized by different degrees. $C_t$ is defined in Eq. \ref{eq:tiouRecallct}. $O_t$ is defined in Eq. \ref{eq:tiouPrecisionOt}.
}\label{fig:qualitative}
\end{figure*}

\subsection{Tightness-aware Intersection-over-Union (TIoU) Metric}
The primary goal of text detection evaluation metrics is to quantify the performance of different methods. Because the major function of detecting text regions is to recognize text, it is a strong requirement for detecting bounding boxes to preserve the completeness of text information and avoid interference with other text instances. However, previous evaluation metrics mentioned in Section 2 do not consider the impact of cutting GT regions and outlier-GT. Meanwhile, it is not easy to distinguish the tightness of detections. Hence, the merits of the detecting methods cannot be thoroughly embodied. To address these problems, our solution is derived from three basic annotating concepts: (a) Annotation does not cut the text instance; (b) annotation contains less background noise, especially outlier text instances; (c) annotations may not be perfect matching the text instance, they should be as much perfect as possible.

\subsubsection{TIoU-Recall}
Intuitively, one GT rectangle $G_i$ cut by a detection bounding box $D_j$ may result in incorrect recognition. Although traditional IoU metrics can measure the tightness between $G_i$ and $D_j$, it can not goal-oriented evaluate the cases shown in Fig. \ref{fig:intro1} (a) and (b) (Detections in (a) and (b) have the same value of IoU (0.66) against the ground truth, while the former one do not recall a few characters of the GT). To solve this issue, the cutting behavior can be penalized by the proportion of intersection in GT.

Firstly, we define the not-recalled area of $G_i$ as $C_t$:
\begin{equation}\label{eq:tiouRecallct}
  C_t = A(G_i)-A(D_j\cap G_i), C_t\in [0,A(G_i)],
\end{equation}
where $A(*)$ means the area of the region. Then, the proportion of intersection in $G_i$ is given by:
\begin{equation}\label{eq:tiouRecallfct}
  f(C_t) = 1-x, x = \frac{C_t}{A(G_i)}.
\end{equation}
Therefore, the final TIoU-Recall is defined as follows:
\begin{equation}\label{eq:tiouRecall}
  TIoU_{Recall} = \frac{A(G_i\cap D_j)*f(C_{t})}{A(G_i\cup D_j)}.
\end{equation}
Equation~\ref{eq:tiouRecall} is a simple but effective solution for managing the cutting behavior, e.g., the TIoU-Recall in Fig. \ref{fig:intro1} (a) and (b) are 0.424 and 0.66, respectively, which implies that missing characters account for a 35.8\% decline in recall.

\subsubsection{TIoU-Precision}
On the other hand, one detection covers several GTs may also affect the recognition results because it is intricate for recognition methods to distinguish which text is target GT, as shown in Fig. \ref{fig:intro1} (c). The proposed solution is to penalize such type of detections to make detection compact for avoiding outlier-GTs. Nevertheless, if the outlier-GTs are inside the target GT region, even the perfect detection bounding box cannot avoid containing these outliers. Therefore, only the outlier-GT region that is inside the detection bounding box but outside the target GT region would be penalized. The area ($O_t$) of the union of all eligible outlier-GTs is calculated using equation~\ref{eq:tiouPrecisionOt}:
\begin{equation}\label{eq:tiouPrecisionOt}
\begin{split}
  O_{t_{ij}} = &A((G_{1} \cap D_j - G_{1}\cap D_j\cap G_i)\cup \\
  &... \cup (G_{i-1} \cap D_j - G_{i-1}\cap D_j\cap G_i)\cup \\
  & (G_{i+1} \cap D_j - G_{i+1}\cap D_j\cap G_i)\cup
  ...\  \cup \\
  & (G_{n} \cap D_j - G_{n}\cap D_j\cap G_i)), \\
  & O_{t_{ij}}\in [0,A(D_j-D_j\cap G_i)].
\end{split}
\end{equation}
Note that for each $G_n (n\neq i)$ that does not intersect with $D_j$, it can be simply ignored, which can improve computing efficiency. Then, the proportion of intersection in $D_j$ is given by:
\begin{equation}\label{eq:tiouPrecisionfot}
  f(O_t) = 1-x, x = \frac{O_t}{A(D_j)}.
\end{equation}
Using equation~\ref{eq:tiouPrecisionfot}, we can define the TIoU-Precision in the same way as TIoU-Recall, as shown in equation~\ref{eq:tiouPrecision}:
\begin{equation}\label{eq:tiouPrecision}
  TIoU_{Precision} = \frac{A(D_j\cap G_i)*f(O_{t})}{A(D_j\cup G_i)}.
\end{equation}

For intuitively understanding the performance of TIoU metrics, we show many examples in Fig. \ref{fig:qualitative}, which will be further discussed in Section 4.2.

\subsection{Tightness-aware Metric}
To calculate the final score, the harmonic mean of recall and precision is usually adopted as the primary metric:
\begin{equation}\label{eq:fmeasure}
  Hmean = 2\frac{Recall\cdot Precision}{Recall+Precision},
\end{equation}
where recall and precision are calculated by
\begin{equation}\label{eq:recallori}
  Recall_{ori} = \frac{\sum Match_{gt_i}}{Num_{gt}},
\end{equation}
\begin{equation}\label{eq:precisionori}
  Precision_{ori} = \frac{\sum Match_{dt_j}}{Num_{dt}}.
\end{equation}
IC13 and IC15 both use a binary [0, 100\%] results to determine $Match_{gt_i}$ and $Match_{dt_j}$.

Binary results can not quantify the tightness of the methods, e.g., two detections $D_1$ and $D_2$ would be regarded the same event if the IoU of $D_1$ is 51\% and that of $D_2$ is 100\%. This is unreasonable because a good evaluation protocol should be able to reflect the discrepancy of detecting tightness.

To this end, we follow and improve IC03 method that directly uses the matching value to represent the score of each item, i.e., binary [0, 100\%] results are replaced by a continuous [0-100\%] index.
However, unlike IC03, TIoU guarantees that all GT can be matched only once. In addition, we use a more reasonable TIoU value as the score, instead of using Eq. \ref{eq:ic03RecallBestMatch} and \ref{eq:ic03PrecisionBestMatch}. Meanwhile, if the highest IoU is less than 0.5, the score is set to 0. Therefore, equations~\ref{eq:recallori} and~\ref{eq:precisionori} become
\begin{equation}\label{eq:siouRecall}
   Recall_{SIoU} = \frac{\sum IoU_{i}}{Num_{gt}},
\end{equation}
\begin{equation}\label{eq:siouPrecision}
   Precision_{SIoU} = \frac{\sum IoU_{j}}{Num_{dt}}.
\end{equation}
Equations~\ref{eq:siouRecall} and~\ref{eq:siouPrecision} are referred to as the {\bf {Score Intersection-over-Union (SIoU)}} metric, which is now tightness-aware for evaluating detection methods. To further consider cutting and outlier intervention, we can use the score of TIoU-Recall and TIoU-Precision to calculate the results, as shown in equations~\ref{eq:sumTiouRecall} and~\ref{eq:sumTiouPrecision}:
\begin{equation}\label{eq:sumTiouRecall}
  Recall_{TIoU} = \frac{\sum TIoU_{recall}}{Num_{gt}}
\end{equation}
\begin{equation}\label{eq:sumTiouPrecision}
   Precision_{TIoU} = \frac{\sum TIoU_{precision}}{Num_{dt}}
\end{equation}

\subsection{The Solution of One-to-many and Many-to-one Metrics.}
\begin{figure}[htb]
\centering\centerline{\includegraphics[width = 8.4cm, height = 4.5cm]{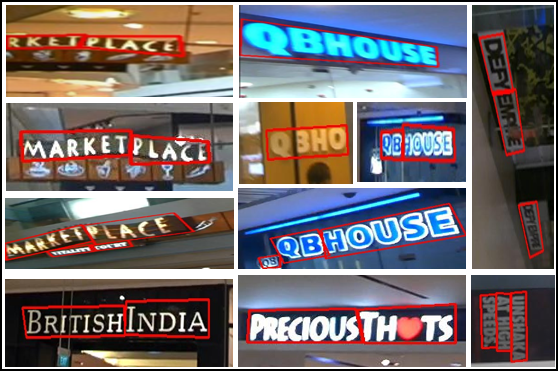}}
\caption{{Examples of annotation inconsistency of GT images in the ICDAR 2015 challenge 4 dataset. Some words without apparent interval are separately annotated while some are annotated with single bounding box.}}\label{fig:annotation_inconsistency}
\end{figure}

Based on our observation, the annotation inconsistency exists on nearly all benchmark datasets. Currently, there is no strict objective annotation protocol, and thus, such ambiguous inconsistency is hard to avoid. Detection methods cannot be fairly evaluated if the GT annotation is inconsistent. Taking the dataset of ICDAR 2015 challenge 4 as an example: This dataset adopts word-level annotation, but sometimes different words are stacked together even without obvious interval. Some examples are shown in Fig. \ref{fig:annotation_inconsistency}. Mandatory labeling the text by words is practically demanding for detection methods. 

The motivation of the OM and MO metrics is to solve the inconsistency of annotation granularity because both word-level and text-line detections may appear simultaneously, and it is hard to judge which one is better for recognition.

Section 2 has already pointed out the weaknesses of the existing OM and MO methods. This section presents a very straightforward solution, i.e., the metric should be used with both word-level and text-line annotations being provided. 
However, to avoid redundant calculation of the same GT, the evaluation requires thorough consideration. Taking the IC15 dataset as an example, the evaluating steps can be outlined as follows (algorithm procedures can be found on appendix):
\begin{itemize}
   \item Creating text-line annotations based on word-level annotations. Each text-line annotation contains at least two word annotations. All the ``don't cared'' regions are ignored in the former. Because IC15 test set contains only 2077 GT boxes, the entire creation procedure is very fast.
   \item The auxiliary text-line annotations are evaluated in advance with the same metric as that used for evaluating word-level annotation. If a detection is matched to a text-line GT, we can calculate the TIoU-precision according to Eq. \ref{eq:tiouRecall}. In the subsequent word-level evaluation stage, the matched detection is ignored. Then, we use word-level annotation and Eq. \ref{eq:13precisionTR} to decide whether the word-level GTs inside this text-line GT are sufficiently recalled. If a word-level GT is recalled, the text-line TIoU-Recall is given by:
   \begin{equation}\label{eq:text-line_tiouRecall}
      TIoU^{*}_{Recall} = \frac{A(G_j\cap D_i)*f(C_{t})}{A(G_j)}. 
   \end{equation}
   In the next word-level evaluating stage, this GT would be considered as ``don't cared''. If the GT is not sufficiently recalled, the TIoU-Recall of this GT will be calculated in word-level stage.   
\end{itemize}


\begin{table*}[!t]
\caption{{Comparison of evaluation methods on ICDAR 2013 for general detection frameworks and previous state-of-the-art methods. $det$: DetEval. $i$: IoU. $e1$: End-to-end recognition results by using CRNN~\cite{shi2017end}. $e2$: End-to-end recognition results by using RARE~\cite{shi2016robust}. t: TIoU.}}
\label{tab:ic13}
\centering
\newcommand{\tabincell}[2]{\begin{tabular}{@{}#1@{}}#2\end{tabular}}
\tiny
\renewcommand\arraystretch{1.6}
\begin{tabular}{|c|ccc|ccc||ccc|ccc||ccc|}
\hline
Methods & $R_{det}$ & $P_{det}$  & $F_{det}$  & $R_{i}$ & $P_{i}$  & $F_{i}$ & $R_{e1}$ & $P_{e1}$  & $F_{e1}$ & $R_{e2}$ & ${P_{e2}}$ & ${F_{e2}}$ & $R_{t}$ & ${P_{t}}$ & ${F_{t}}$ \\
\hline 
Faster R-CNN (VGG16) \cite{ren2015faster}  &  0.410  & 0.549  & 0.469 &  0.615  & 0.752  & 0.676 & 0.396 & 0.432  & 0.413 & 0.406 & 0.442 & 0.423 & 0.377 & 0.554 & 0.448 \\
SSD (300x300) \cite{liu2016ssd}  &  0.476  & 0.88  & 0.618 &  0.484  & 0.886  & 0.626 &  0.398  & 0.639  & 0.491 & 0.391 & 0.629 & 0.483 & 0.377 & 0.727 & 0.496 \\
YOLO-v2 (320x320) \cite{redmon2016yolo9000}  &  0.431  & 0.772  & 0.553 &  0.481  & 0.877  & 0.621 &  0.372  & 0.548  & 0.443 & 0.526 & 0.571 & 0.547 & 0.339 & 0.682 & 0.453 \\
YOLO-v3 (320x320) \cite{redmon2018yolov3}  &  0.648  & 0.823  & 0.725 &  0.68  & 0.874  & 0.765 &  0.519  & 0.611  & 0.561 & 0.523 & 0.516 & 0.566 & 0.502 & 0.696 & 0.583 \\
YOLO-v3 (512x512) \cite{redmon2018yolov3}  &  0.694  & 0.867  & 0.771 &  0.721  & 0.895  & 0.799 &  0.566  & 0.65  & 0.605 & 0.585 & 0.672 & 0.625 & 0.549 & 0.73 & 0.627 \\
Mask R-CNN \cite{He2017Mask}  &  0.767  & 0.793  & 0.780 &  0.718  & 0.715  & 0.716 &  0.544  & 0.494  & 0.518 & 0.58 & 0.525 & 0.551 & 0.527 & 0.545 & 0.536 \\
R-FCN (resNet-50) \cite{dai2016r} &  0.603  & 0.796  & 0.686 &  0.656  & 0.869  & 0.748 &  0.527 & 0.627  & 0.573 & 0.543 & 0.647 & 0.59 & 0.488 & 0.712 & 0.579 \\
Faster R-CNN-FPN \cite{lin2018focal} &  0.674  & 0.882  & 0.764 &  0.686  & 0.875  & 0.769 &  0.578  & 0.678  & 0.624 & 0.597 & 0.699 & 0.644 & 0.551 & 0.737 & 0.631 \\
RetinaNet (resNet-50-FPN) \cite{lin2018focal} &  0.452  & 0.901  & 0.602 &  0.46  & 0.906  & 0.611 &  0.409  & 0.744  & 0.528 & 0.385 & 0.7 & 0.497 & 0.375 & 0.77 & 0.504 \\
\hline
East~\cite{zhou2017east} & 0.707  & 0.816  & 0.758 & 0.731  & 0.835  & 0.779 & 0.588  & 0.595  & 0.591 & 0.6 & 0.607 & 0.603 & 0.567 & 0.684 & 0.620 \\
SegLink~\cite{shi2017detecting}  &  0.6  & 0.739  & 0.662 &  0.572  & 0.666  & 0.615 &  0.485  & 0.497  & 0.491 & 0.495 & 0.507 & 0.501 & 0.387 & 0.471 & 0.425 \\
PixelLink \cite{deng2018pixellink}  &  0.633  & 0.679  & 0.655 &  0.621  & 0.618  & 0.619 &  0.539  & 0.481  & 0.508 & 0.549 & 0.489 & 0.517 & 0.432 & 0.442 & 0.437 \\
TextBox~\cite{liao2017textboxes}  &  0.731  & 0.896  & 0.805 &  0.741  & 0.892  & 0.809 & 0.594  & 0.643  & 0.618 & 0.614 & 0.664 & 0.638 & 0.564 & 0.712 & 0.629 \\
SWT-MSER \cite{epshtein2010detecting,matas2004robust}  &  0.371  & 0.258  & 0.305 & 0.17  & 0.181  & 0.175 &  0.083  & 0.075  & 0.079 & 0.317 & 0.243 & 0.275 & 0.122 & 0.136 & 0.129 \\
FEN~\cite{zhang2017feature}  &  0.899  & 0.947  & 0.923 &  0.885  & 0.934  & 0.909 &  0.719  & 0.716  & 0.717 & 0.759 & 0.757 & 0.758 & 0.721 & 0.783 & 0.751 \\
R2CNN~\cite{jiang2017r2cnn}  &  0.905  & 0.943  & 0.923 &  0.875  & 0.908  & 0.891 &  0.745  & 0.732  & 0.738 & 0.762 & 0.749 & 0.756 & 0.687 & 0.721 & 0.704 \\
MaskTextSpotter \cite{lyu2018mask}  &  0.886  & 0.95  & 0.917 &  0.873  & 0.935  & 0.903 &  0.751  & 0.752  & 0.752 & 0.766 & 0.766 & 0.766 & 0.733 & 0.809 & 0.769 \\
WordSup~\cite{hu2017wordsup}  &  0.871  & 0.928  & 0.899 &  0.702  & 0.821  & 0.757 &  0.611  & 0.648  & 0.629 & 0.624 & 0.662 & 0.642 & 0.533 & 0.626 & 0.575 \\
AF-RPN~\cite{zhong2018anchor}  &  0.896  & 0.945  & 0.92 &  0.854  & 0.902  & 0.877 &  0.731  & 0.72  & 0.725 & 0.756 & 0.744 & 0.75 & 0.665 & 0.711 & 0.687 \\
\hline

\end{tabular}
\end{table*}

\begin{table*}[!t]
\caption{{Comparison of metrics on the ICDAR 2015 challenge 4. Word\&Text-Line Annotations use our new solution to address OM and MO issues. i: IoU. s: SIoU. t: TIoU.}}
\label{tab:ic15}
\centering
\scriptsize
\renewcommand\arraystretch{1.6}
\begin{tabular}{|c|ccc|ccc|ccc||ccc|ccc|}
\hline
\multirow{2}*{Methods} & \multicolumn{9}{c||}{Original Word-level-Only Annotations} &  \multicolumn{6}{c|}{Word\&Text-Line Annotations}\\
\cline{2-4}
\cline{5-7}
\cline{8-10}
\cline{11-13}
\cline{14-16}
      & $R_{i}$ & $P_{i}$  & $F_{i}$ & $R_{s}$ & ${P_{s}}$ & ${F_{s}}$ & $R_{t}$ & ${P_{t}}$ & ${F_{t}}$ & $R_{i}$ & $P_{i}$  & $F_{i}$ & $R_{t}$ & ${P_{t}}$ & ${F_{t}}$ \\
\hline
SegLink~\cite{shi2017detecting} &  0.728  & 0.802  & 0.764 & 0.54 & 0.594 & 0.566 & 0.467 & 0.581 & 0.517 & 0.747 & 0.836 & 0.789 & 0.505 & 0.598 & 0.548 \\
East~\cite{zhou2017east} &  0.772  & 0.846  & 0.808 & 0.593 & 0.65 & 0.62 & 0.528 & 0.635 & 0.576 & 0.785 & 0.864 & 0.823 & 0.567 & 0.64 & 0.601 \\
RRD~\cite{liao2018rotation} &  0.778  & 0.868  & 0.821 & 0.594 & 0.663 & 0.627 & 0.515 & 0.652 & 0.575 & 0.783 & 0.879 & 0.829 & 0.53 & 0.653 & 0.585 \\
PixelLink~\cite{deng2018pixellink} &  0.817  & 0.829  & 0.823 & 0.616 & 0.626 & 0.621 & 0.552 & 0.618 & 0.583 & 0.829 & 0.851 & 0.84 & 0.585 & 0.627 & 0.605 \\
TextBox++~\cite{liao2018textboxes++} &  0.808  & 0.891  & 0.847 & 0.619 & 0.683 & 0.649 & 0.537 & 0.672 & 0.597 & 0.812 & 0.9 & 0.854 & 0.549 & 0.67 & 0.603 \\
DMPNet~\cite{liu2017deep} &  0.765  & 0.757  & 0.761 & 0.564 & 0.558 & 0.561 & 0.479 & 0.546 & 0.51 & 0.781 & 0.779 & 0.78 & 0.512 & 0.554 & 0.532 \\
WordSup~\cite{hu2017wordsup} &  0.773  & 0.805  & 0.789 & 0.568 & 0.591 & 0.579 & 0.49 & 0.577 & 0.53 & 0.785 & 0.831 & 0.807 & 0.522 & 0.588 & 0.553 \\
R2CNN~\cite{jiang2017r2cnn} &  0.828  & 0.887  & 0.855 & 0.641 & 0.687 & 0.663 & 0.559 & 0.676 & 0.612 & 0.831 & 0.901 & 0.865 & 0.577 & 0.676 & 0.622 \\
AF-RPN~\cite{zhong2018anchor} & 0.832 & 0.891  & 0.861 & 0.645 & 0.69 & 0.667 & 0.577 & 0.677 & 0.623 & 0.844 & 0.912 & 0.877 & 0.607 & 0.681 & 0.642 \\
MaskTextSpotter~\cite{lyu2018mask} &  0.795  & 0.89  & 0.84 & 0.6 & 0.671 & 0.633 & 0.527 & 0.658 & 0.585 & 0.803 & 0.906 & 0.851 & 0.549 & 0.662 & 0.6 \\
\hline
\end{tabular}
\end{table*}

\begin{figure}[htb]
\begin{minipage}[b]{.32\linewidth}
  \centering
  \centerline{\includegraphics[width=2.8cm, height = 3.7cm]{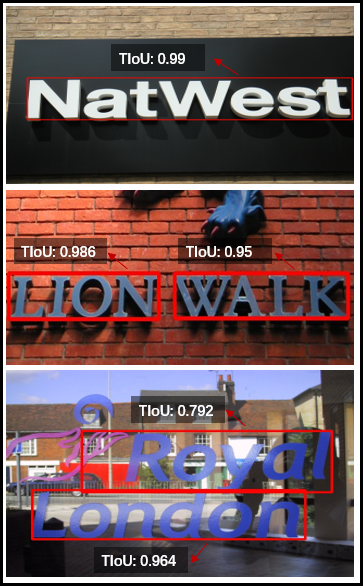}}
  \centerline{(a) MTS.}\medskip
\end{minipage}
\hfill
\begin{minipage}[b]{0.32\linewidth}
  \centering
  \centerline{\includegraphics[width=2.8cm, height = 3.7cm]{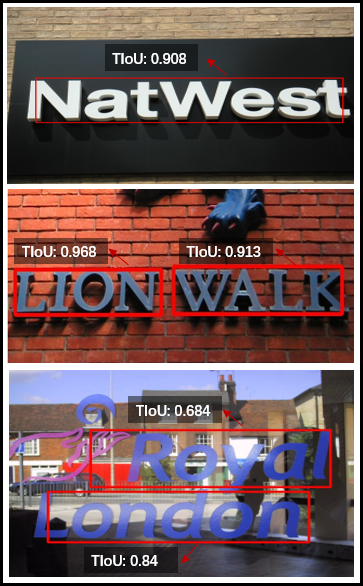}}
  \centerline{(b) FEN.}\medskip
\end{minipage}
\hfill
\begin{minipage}[b]{0.32\linewidth}
  \centering
  \centerline{\includegraphics[width=2.8cm, height = 3.7cm]{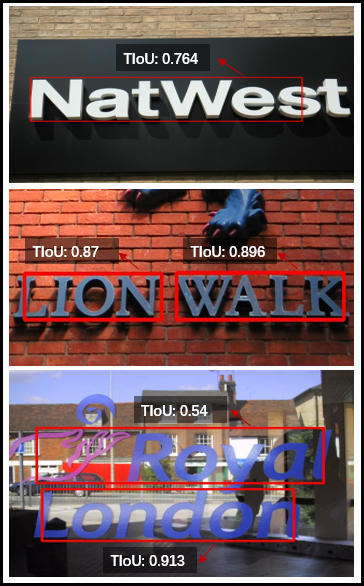}}
  \centerline{(c) R2CNN.}\medskip
\end{minipage}
\caption{According to Table \ref{tab:ic13}, the performances of these three methods are comparable under previous metrics but different while using TIoU metric. This is because TIoU can perceive the completeness of the target GT, the compactness of the detection, and the tightness of the matching. We can also find that a detection that achieve 0.9 or higher TIoU value is nearly perfect, which could be directly serve as a . 
MTS: MaskTextSpotter.}
\label{fig:res_13Tighter}
\end{figure}

\begin{figure}[htb]
\begin{minipage}[b]{.32\linewidth}
  \centering
  \centerline{\includegraphics[width=2.8cm, height = 3.7cm]{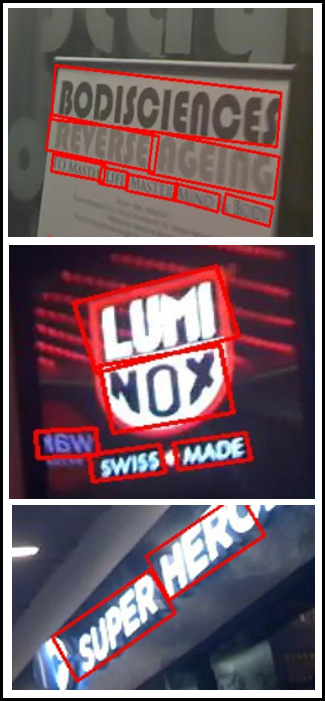}}
  \centerline{(a) East.}\medskip
\end{minipage}
\hfill
\begin{minipage}[b]{0.32\linewidth}
  \centering
  \centerline{\includegraphics[width=2.8cm, height = 3.7cm]{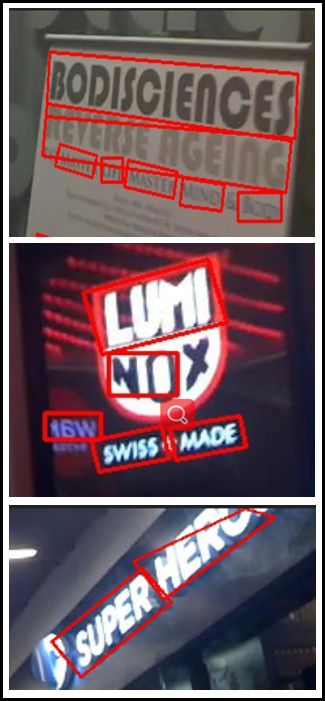}}
  \centerline{(b) PixelLink.}\medskip
\end{minipage}
\hfill
\begin{minipage}[b]{0.32\linewidth}
  \centering
  \centerline{\includegraphics[width=2.8cm, height = 3.7cm]{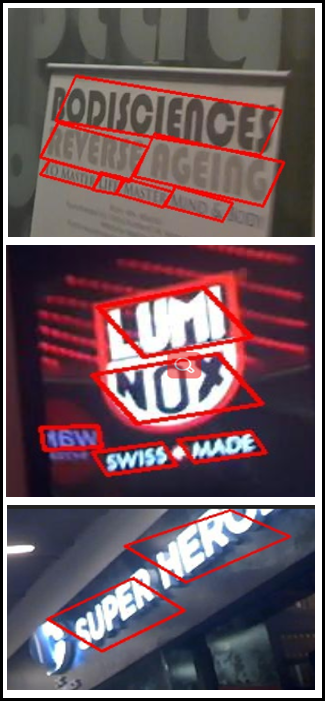}}
  \centerline{(c) RRD.}\medskip
\end{minipage}
\caption{The visualization results are corresponding to Table \ref{tab:ic15}. The performances of PixelLink and RRD are much better than East under previous metrics. However, in TIoU metric, the results are comparable, and East can even outperform RRD, which is mainly because the detections of East are tighter than other two methods.}
\label{fig:res_15Tighter}
\end{figure}

\section{Experiments}
This section quantitatively and qualitatively tested the proposed TIoU metric on two most popular scene text detection datasets - IC13 \cite{Karatzas2013ICDAR} and IC15 \cite{karatzas2015icdar}.

\subsection{Quantitative evaluation}
Quantitative experiments have been conducted to evaluate the differences between the previous metrics and the proposed TIoU metric. To evaluate the results, several state-of-the-art methods were adopted. All the methods were selected based on three criteria: 1) The methods have been frequently referred to in the literature; 2) the authors are willing to provide their detection results for our evaluation; 3) the authors or other researchers have published the source code or test model, thus making the result easy to reproduce, such as PixelLink \cite{deng2018pixellink}. Some results of IC13 directly used the model of IC15 provided by the authors (because they did not provide IC13 model).  

In addition, we further compared the metrics by several popular general object detection frameworks on the IC13 dataset. All the training data for these general detection frameworks are strictly the same, including 1715 samples from FORU \cite{scut-foru} and 229 samples from the official IC13 training set.

To further demonstrate the effectiveness of the TIoU metric, we adopted two recognition methods (CRNN \cite{shi2017end} and RARE \cite{shi2016robust}) to evaluate end-to-end results. The end-to-end metric assesses the localization in the same way as detection task, and then evaluate the recall, precision, and Hmean based on transcription perfect match \cite{karatzas2015icdar}.

The results are shown in Table \ref{tab:ic13} and Table \ref{tab:ic15}, respectively.
From Table \ref{tab:ic13}, several new prospects can be observed: 
\begin{itemize}
  \item {\bf {The values of TIoU show a similar tendency as end-to-end evaluation results}}. For example, Faster R-CNN outperforms SSD by 5\% in the original IoU metric (0.676 vs. 0.626); however, in TIoU metric, the latter surpasses the former by almost 5\%, which is corresponding to End-to-end results: For $F_{e1}$ and $F_{e2}$, the latter exceeds the former by 7.8\% and 6\%, respectively.
  To make it clearer, we further draw a line chart which can be more intuitive to visualize the correlations, as shown in Figure \ref{fig:correlations}. From the figure, we can find the results of the $F_{e1}$, $F_{e2}$, and $F_{TIoU}$ are extremely similar whereas the results of the $F_{IoU}$ are clearly different from the other three results. On the other hand, the correlation coefficients $r$($F_{TIoU}$ ; $F_{e1}$) and $r$($F_{TIoU}$ ; $F_{e2}$) are higher than $r$($F_{IoU}$ ; $F_{e1}$) and $r$($F_{IoU}$ ; $F_{e2}$), with 0.5\% and 2.6\%, respectively.
  \begin{figure}[htb]
  \begin{minipage}[b]{0.48\linewidth}
    \centering
    \centerline{\includegraphics[width=3.2cm, height = 2cm]{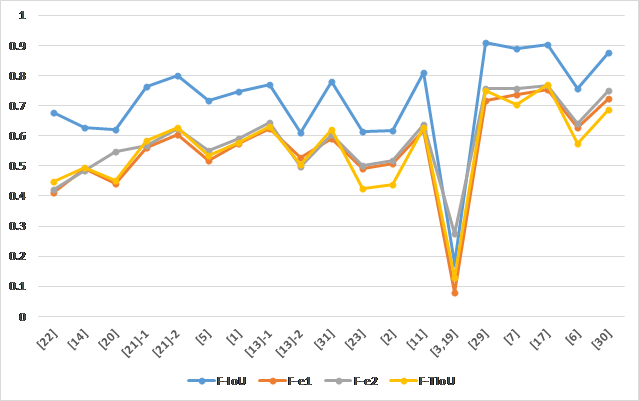}}
    \centerline{(a) Line chart.}\medskip
  \end{minipage}
  \hfill
  \begin{minipage}[b]{0.48\linewidth}
    \centering
    \centerline{\includegraphics[width=4.4cm, height = 2cm]{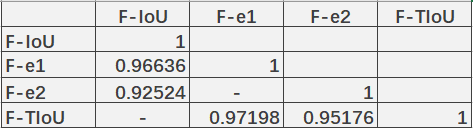}}
    \centerline{(b) Correlation coefficients.}\medskip
  \end{minipage}
  \caption{(a) X-axis represents the detection methods listed in Table \ref{tab:ic13}, and Y-axis represents the values of the F-measures. (Zoom in for better visualization)}
  \label{fig:correlations}
  \end{figure}
  
  \item {\bf {TIoU is tightness-aware}}. FEN, R2CNN, and MaskTextSpotter (MTS) are comparable under previous metrics; however, in TIoU metrics, the results are much different, with 0.751, 0.704, and 0.769, respectively for these three methods. This is mainly because the TIoU is tightness-aware: Examples are shown in Fig. \ref{fig:res_13Tighter}, MTS is the tightest method, and next is FEN and R2CNN, which are corresponding to the TIoU ranks.

  \item {\bf {Text detection is still a challenging task}}. Comparing the TIoU metric and previous metrics among all detection methods, most of the results consistently decrease by more than 10\%, which indicates there are still insufficiencies of text detection methods. In addition, even the best performance of TIoU Hmean is under 0.8, which further shows that designing a robust text detection method is not an easy task.
\end{itemize}

\begin{figure}[htb]
\begin{minipage}[b]{.32\linewidth}
  \centering
  \centerline{\includegraphics[width=2.8cm, height = 4.8cm]{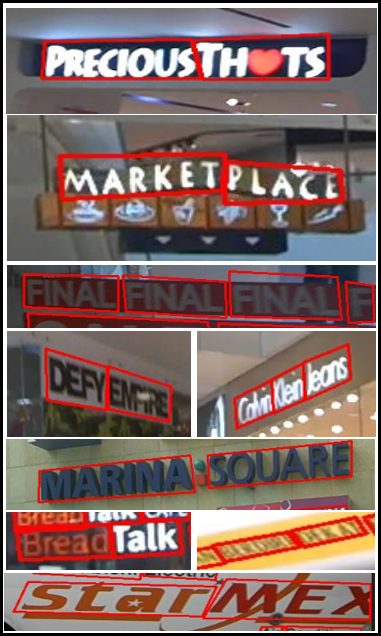}}
  \centerline{(a) Original GT.}\medskip
\end{minipage}
\hfill
\begin{minipage}[b]{0.32\linewidth}
  \centering
  \centerline{\includegraphics[width=2.8cm, height = 4.8cm]{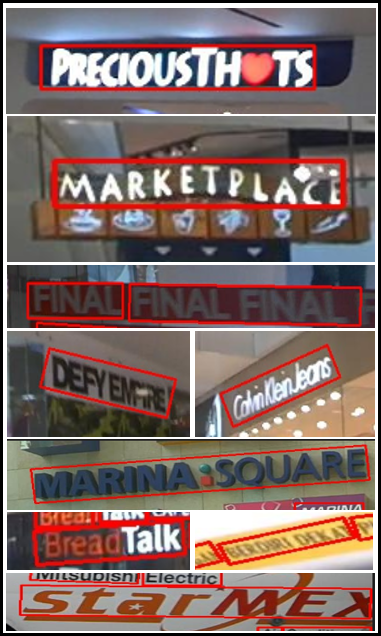}}
  \centerline{(b) PixelLink.}\medskip
\end{minipage}
\hfill
\begin{minipage}[b]{0.32\linewidth}
  \centering
  \centerline{\includegraphics[width=2.8cm, height = 4.8cm]{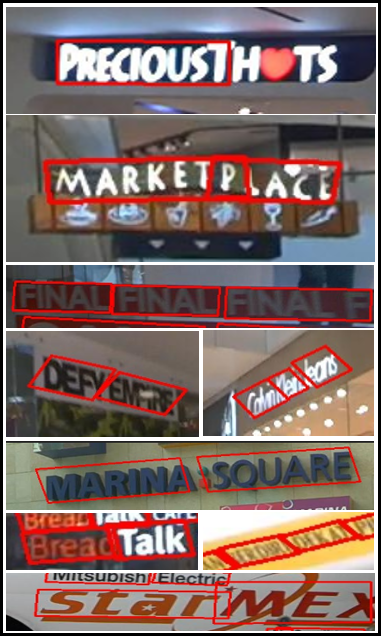}}
  \centerline{(c) TextBox++.}\medskip
\end{minipage}
\caption{Examples of ambiguous detections by different methods. If using original metrics, detections of PixelLink in this figure are false positives, which is the main reason that PixelLink (0.823) is worse than TextBox++ (0.847) (refer to Tab. \ref{tab:ic15}). While using our joint word\&text-line solution, such text-line detections can be fairly evaluated, and PixelLink (0.605) can even outperform TextBox++ (0.603) in TIoU metric (refer to Tab. \ref{tab:ic15}).}
\label{fig:res_textlineDemonstration}
\end{figure}

Experiments on the IC15 dataset also demonstrate the effectiveness of the proposed metrics, as shown in Table \ref{tab:ic15}:
\begin{itemize}
  \item {\bf {TIoU is also tightness-aware on multi-oriented dataset}}. For example, PixelLink and RRD outperform East in previous IoU metric $F_i$; however, the gap reduces using SIoU $F_s$, which indicates that the detections of East are tighter; in TIoU metric $F_t$, East can further surpass RRD. This is because TIoU metric can reflect the tightness of different detection methods. As some detection results shown in Fig. \ref{fig:res_15Tighter}, although the rotated detecting rectangles of RRD satisfy the requirement of previous IoU metric, they may cut the GT regions to some extent, which is the main reason that cause the decrease of TIoU. This example also implies that tighter quadrilateral detection methods may reasonably be more benefit from TIoU metric than rectangle-based methods.
  \item {\bf {Objectively evaluated Text-line detections}}. Because there are many ambiguous annotation granularities on the IC15 dataset (shown in Fig. \ref{fig:annotation_inconsistency}), we have added the text-line level annotations of this dataset, and using our new solution (details in Section 3.3) for OM and MO matching. The results show that the results of all methods can be increased by different degrees in both the IoU and the TIoU metrics. This is mainly because some fine text line detections are fairly evaluated instead of treating them as false positives. For example, before using our new Joint-word\&text-line evaluation metric, TextBox++ outperformed PixelLink for a large margin (2.4\%). However, the former is worse than the latter by 0.2\% with joint text-line annotations. The variation can be explained by visualization results shown in Fig. \ref{fig:res_textlineDemonstration}. The quantitative and qualitative results both demonstrate that our solution can reveal the potential novelty of segmentation-based methods. 

\end{itemize}

\subsection{Qualitative evaluation}
Fig. \ref{fig:qualitative} shows some examples of the qualitatively evaluated results. If previous IoU metrics are used, all detections and target GTs of Fig. \ref{fig:qualitative} are considered to have 100\% precision and recall because the their IoUs are all higher than 0.5. Clearly, this is unreasonable because some detections significantly reduce the recognition performance, such as Detection \#1 in Fig. \ref{fig:qualitative} (a) and Detection \#3 in Fig. \ref{fig:qualitative} (b).

Using the TIoU metric can assist in avoid these phenomenons. As shown in Fig. \ref{fig:qualitative}, the TIoU recall and precision are associated to recognition to some extent. For example, NED of detection \#1 in (a) is 0.625, which means only 37.5\% of the text can be correctly recognized because of the cutting effect. In this case, the TIoU-recall is also equal to 37.5\%. This coincidence doesn't imply there is a specific relationship between the TIoU and recognition performance because NED can vary depending on the recognition methods. However, in Fig. \ref{fig:qualitative}, it is intuitive that high NED cases usually have a relative strong penalizes of TIoU value. Note that, for some high IoU (but not perfect detections) such as Detection \#2 in Fig. \ref{fig:qualitative} (a), the recall only decreases slightly (from 0.891 to 0.866), which still meets our subjective judgment.


\section{Conclusion}
In this paper, we presented a goal-oriented TIoU metric to address many drawbacks of previous metrics. The TIoU metric is simple but intuitive, which moderately consider two common detection behaviors that may significantly influence the recognition. It uses the TIoU score as a measure of recall and precision to perceive the tightness of detection methods. Quantitative experiments on ICDAR 2013 and ICDAR 2015 datasets showed that the proposed metric has a similar tendency of the end-to-end detection and recognition results among general object detection frameworks and previous state-of-the-art text detection methods. Qualitative experiments further demonstrated the reduction from previous IoU to TIoU is mainly because TIoU can perceive and quantify the tightness of detections.

In addition, because the previous one-to-many and many-to-one metrics have many drawbacks, we proposed a straightforward solution to solve these issues. The results on the ICDAR 2015 dataset show that all methods can be improved with different degrees using our new solution. This is mainly because the proposed method can reasonably evaluate the text-line detections instead of roughly regarding them as false positives.

In future, we will try to use TIoU metric to guide training because its characteristics may be benefited to provide a strong supervision. In addition, it can also be used to help incremental or semi-supervised learning because TIoU can judge whether a detection is suitable to serve as a new GT annotation.

\section*{Acknowledgements}
This research is supported in part by GD-NSF (no. 2017A030312006), the National Key Research and Development Program of China (No. 2016YFB1001405), NSFC (Grant No.: 61673182, 61771199), and GDSTP (Grant No.:2017A010101027), GZSTP(no. 201704020134).

{\small
\bibliographystyle{ieee}
\bibliography{EvalMetric}
}

\newpage
\section*{Appendix}

\noindent 
{\bf Pseudo code of OM \& MO solution.} Algorithm \ref{alg:joint} summarizes the joint Word\&Text-Line annotation evaluating procedure. Note that not all details are covered by \ref{alg:joint}, which can be found on the source code.  \\

\begin{algorithm}[htb]  
\caption{ Joint Word\&Text-Line Evaluation.} 
\label{alg:joint}  
\begin{algorithmic}[1]  
\STATE \textbf{Input:} \\
$S$ - Dataset to be evaluated.\\
$W$ - Word-Level annotations. \\
$W_{i}$: the $i$-th annotation of $W$.\\
$T$ - Text-Line annotations.\\
$T_{j}$: the $j$-th annotation of $T$.\\
$D$ - Detection results.\\
$M$ - Matching indicator. All zero in the beginning.\\ 
\STATE \textbf{Evaluation Procedure:} \\
{(\bf a) Creating $T$ from $W$ of $S$ manually.} \\
\ \ \ \ \ \ (i) $T$ ignores all ``don't care'' instances of $W$.\\
\ \ \ \ \ \ (ii) Normally, each annotation of $T$ contains at least \\
\ \ \ \ \ \ \ \ \ \ two instances of $W$.\\

{(\bf b) Using ``don't care'' instances of $W$ to distinguish ``don{}'t care'' detections of $D$.}\\
\ \ \ \ \ \ (i) If $\frac{Area(W_{i}\cap D_{j})}{Area(D_{j})} > 0.5$, $D_{j}$ is marked as ``don't \\
\ \ \ \ \ \ \ \ \ \ care''. \\

{(\bf c) Creating matching indexes between $T$ and $W$.} \\
\ \ \ \ \ \ (i) This step can be done during step ($\bf a$). Deciding \\
\ \ \ \ \ \ \ \ \ \ which $W_{i}$ belongs to which $T_{j}$. \\
\ \ \ \ \ \  (ii) If not (i), then if $\frac{Area(T_{j}\cap W_{i})}{Area(W_{i})} > 0.5$, simply \\
\ \ \ \ \ \ \ \ \ \ marking $W_{i}$ belongs to $T_{j}$.\\

{(\bf d) Evaluating $D$ on $T$ in advance.} \\
\ \ \ \ \ \ If $M_{D_{i}}$ is zero and $D_i$ is not marked as ``don't care':\\ 
\ \ \ \ \ \ \ \ \ \ \ \ \ \ If IoU of $D_i$ and $T_j$ is $> 0.5$: \\
\ \ \ \ \ \ \ \ \ \ \ \ \ \ \ \ \ \ Accumulating TIoU precision \\
\ \ \ \ \ \ \ \ \ \ \ \ \ \ \ \ \ \ $M_{D_{i}}$ $=$ 1 \\
\ \ \ \ \ \ \ \ \ \ \ \ \ \ \ \ \ \ For $W_1$ ... $W_k$ that belong to $T_j$: \\
\ \ \ \ \ \ \ \ \ \ \ \ \ \ \ \ \ \ \ \ \ \ If $\frac{Area(W_{k}\cap D_{i})}{Area(W_{k})} < 0.5$: \\
\ \ \ \ \ \ \ \ \ \ \ \ \ \ \ \ \ \ \ \ \ \ \ \ \ \ leave $W_{k}$ to the step (f) \\
\ \ \ \ \ \ \ \ \ \ \ \ \ \ \ \ \ \ \ \ \ \ else:\\
\ \ \ \ \ \ \ \ \ \ \ \ \ \ \ \ \ \ \ \ \ \ \ \ \ \ Accumulating TIoU recall using Eq. \ref{eq:text-line_tiouRecall} \\ 
\ \ \ \ \ \ \ \ \ \ \ \ \ \ \ \ \ \ \ \ \ \ \ \ Marking $W_{k}$ as ``don't care'' \\
\ \ \ \ \ \ \ \ \ \ \ \ \ \ \ \ \ \  end for \\
\ \ \ \ \ \ \ \ \ \ \ \ \ \ end if \\
\ \ \ \ \ \ end if \\

{(\bf e) Using new ``don't care'' instances of $W$ to distinguish ``don{}'t care'' detections of $D$.}\\
\ \ \ \ \ \ (i) Same as step (b) \\

{(\bf f) Evaluating $D$ on $W$.} \\
\ \ \ \ \ \ If $M_{D_{i}}$, $M_{W_{j}}$ are zero and $D_i$, $W_{j}$ $\neq$  ``don't care':\\ 
\ \ \ \ \ \ \ \ \ \ \ \ \ \ If IoU of $D_i$ and $W_j$ is $> 0.5$: \\
\ \ \ \ \ \ \ \ \ \ \ \ \ \ \ \ \ \ Accumulating TIoU precision \\
\ \ \ \ \ \ \ \ \ \ \ \ \ \ \ \ \ \ Accumulating TIoU recall \\
\ \ \ \ \ \ \ \ \ \ \ \ \ \ \ \ \ \ $M_{D_{i}}$ $=$ 1 \\
\ \ \ \ \ \ \ \ \ \ \ \ \ \ \ \ \ \ $M_{W_{j}}$ $=$ 1 \\
\ \ \ \ \ \ \ \ \ \ \ \ \ \ end if \\
\ \ \ \ \ \ end if \\
\STATE \textbf{Output:}\\
Final TIoU precision. \\
Final TIoU Recall. \\
Using Eq. \ref{eq:fmeasure} to calculate final TIoU Hmean.
\end{algorithmic}
\end{algorithm}

\begin{figure}[htb]
\begin{minipage}[b]{.49\linewidth}
  \centering
  \centerline{\includegraphics[width=4.2cm, height = 6.8cm]{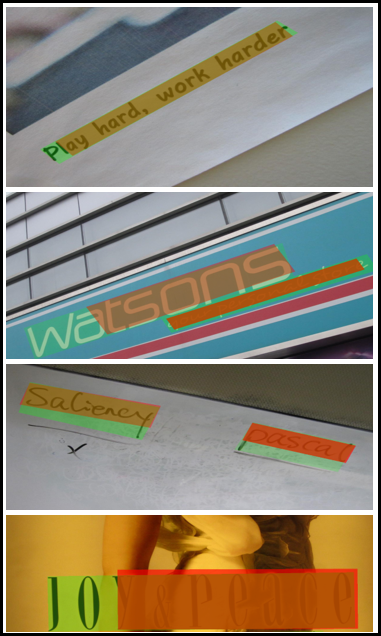}}
  \centerline{(a) Improved Mask R-CNN.}\medskip
\end{minipage}
\hfill
\begin{minipage}[b]{0.49\linewidth}
  \centering
  \centerline{\includegraphics[width=4.2cm, height = 6.8cm]{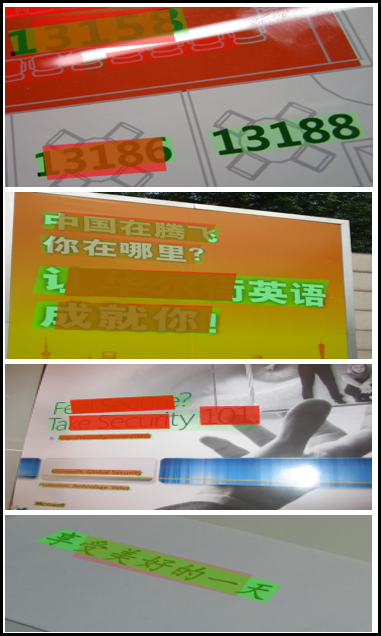}}
  \centerline{(b) East.}\medskip
\end{minipage}
\caption{Visualization results of MSRA-TD500. Green: Ground truth. Red: Detection result. Orange: Overlapping region. Note that previous metrics would regard all these detections with 100\% recall and precision. }
\label{fig:msra}
\end{figure}

\noindent
{\bf Experiments on Non-Latin text.} The experiments on the main paper only conducted on the word-level Latin datasets. Actually, it may be more effective for detecting long text lines. Thus we further supplemented an experiment on well-known MSRA-TD500 \cite{Yao2012Detecting}. Because we cannot retrieve the others' previous detection results of this dataset, we train East \cite{zhou2017east} and an improved version of Mask R-CNN with additional data selected from RCTW-17 \cite{Shi2017ICDAR2017} to test the TIoU metric. The results are shown in Table \ref{tab:msra}, and some of the qualitative results are shown in Figure \ref{fig:msra}. Note that, during the evaluation, we calculate the exact value of the area of the polygon instead of using approximate area calculation in \cite{Yao2012Detecting}. It can be seen from Table \ref{tab:msra} that the recall, precision, and Hmean all drop significantly, which are mainly because there exists many defective detections shown in Figure \ref{fig:msra}. TIoU metric also highlights the difference between object detection and text detection task, showing there still exists a large room for detection methods to improve.

\begin{table}[!h]
\caption{{Comparison of metrics on the TD500. i: IoU. t: TIoU.}}
\label{tab:msra}
\centering
\scriptsize
\renewcommand\arraystretch{1.6}
\begin{tabular}{|c|ccc|ccc|}
\hline
Methods & $R_{i}$ & ${P_{i}}$ & ${F_{i}}$ & $R_{t}$ & ${P_{t}}$ & ${F_{t}}$ \\
\hline
East \cite{zhou2017east} & 0.615  & 0.49  & 0.546 & 0.411  & 0.369  & 0.389 \\
\hline
Mask R-CNN++ \cite{He2017Mask} & 0.832  & 0.837  & 0.834 & 0.638  & 0.679  & 0.658 \\
\hline
\end{tabular}
\end{table}

\end{document}